\documentclass[letterpaper]{article} 
\usepackage{aaai25}  
\usepackage{times}  
\usepackage{helvet}  
\usepackage{courier}  
\usepackage[hyphens]{url}  
\usepackage{graphicx} 
\usepackage{natbib}  
\usepackage{caption} 
\usepackage{algorithm}
\usepackage{algorithmic}

\usepackage{multirow}
\usepackage{subcaption}
\usepackage{enumitem}
\usepackage{amssymb}
\usepackage{amsmath}
\usepackage{colortbl}
\usepackage{booktabs}
\usepackage{marvosym}
\usepackage{fancyhdr}
\usepackage{newfloat}
\usepackage{listings}
\DeclareCaptionStyle{ruled}{labelfont=normalfont,labelsep=colon,strut=off} 
\floatstyle{ruled}
\newfloat{listing}{tb}{lst}{}
\floatname{listing}{Listing}
\title{TdAttenMix: Top-Down Attention Guided Mixup}
\author{
    Zhiming Wang\textsuperscript{\rm 1},
    Lin Gu\textsuperscript{\rm 2, 3},
    Feng Lu\textsuperscript{\rm 1}\textsuperscript{\thanks{Corresponding Author.}}
}
\affiliations{
    \textsuperscript{\rm 1}State Key Laboratory of VR Technology and Systems, School of CSE, Beihang University\\

    \textsuperscript{\rm 2}RIKEN AIP\\
    \textsuperscript{\rm 3}The University of Tokyo, Japan\\
    \{zy2306418,lufeng\}@buaa.edu.cn, lin.gu@riken.jp
}

\begin{document}

\maketitle

\begin{abstract}
CutMix is a data augmentation strategy that cuts and pastes image patches to mixup training data. Existing methods pick either random or salient areas which are often inconsistent to labels, thus misguiding the training model. By our knowledge, we integrate human gaze to guide cutmix for the first time. Since human attention is driven by both high-level recognition and low-level clues, we propose a controllable Top-down Attention Guided Module to obtain a general artificial attention which balances top-down and bottom-up attention. The proposed TdATttenMix then picks the patches and adjust the label mixing ratio that focuses on regions relevant to the current label. Experimental results demonstrate that our TdAttenMix outperforms existing state-of-the-art mixup methods across eight different benchmarks. Additionally, we introduce a new metric based on the human gaze and use this metric to investigate the issue of image-label inconsistency.
\end{abstract}

\begin{links}
\link{Code}{https://github.com/morning12138/TdAttenMix}
\end{links}

\section{Introduction}
Thanks to large amount of data, Deep Neural Networks (DNNs) have achieved significant success in recent years across a variety of applications, including recognition~\cite{vit, Dlme, Cui23ECCV, Tan_2022_CVPR, Chen_2021_ICCV}, graph learning~\cite{xia2022pretraining, 9632431, Physicalattack}, and video processing~\cite{sg_net, Cui_2021_ICCV, liu2021densernet, zhao2022tracking}. However, the data-hungry problem~\cite{vit, datahungry2}  leads to  overfitting when the training data are scarce. Therefore, a series of data augmentation techniques called mixup are proposed to alleviate this issue and  enhance DNNs' generalization capabilities. Among them, CutMix~\cite{cutmix} is an effective strategy that randomly crops a patch from the source image and pastes it into the target image. The label is then mixed by the source and target labels in proportion to the crop area ratio.

\begin{figure}
    \centering
    \includegraphics[width=1\linewidth,height=0.4\linewidth]             {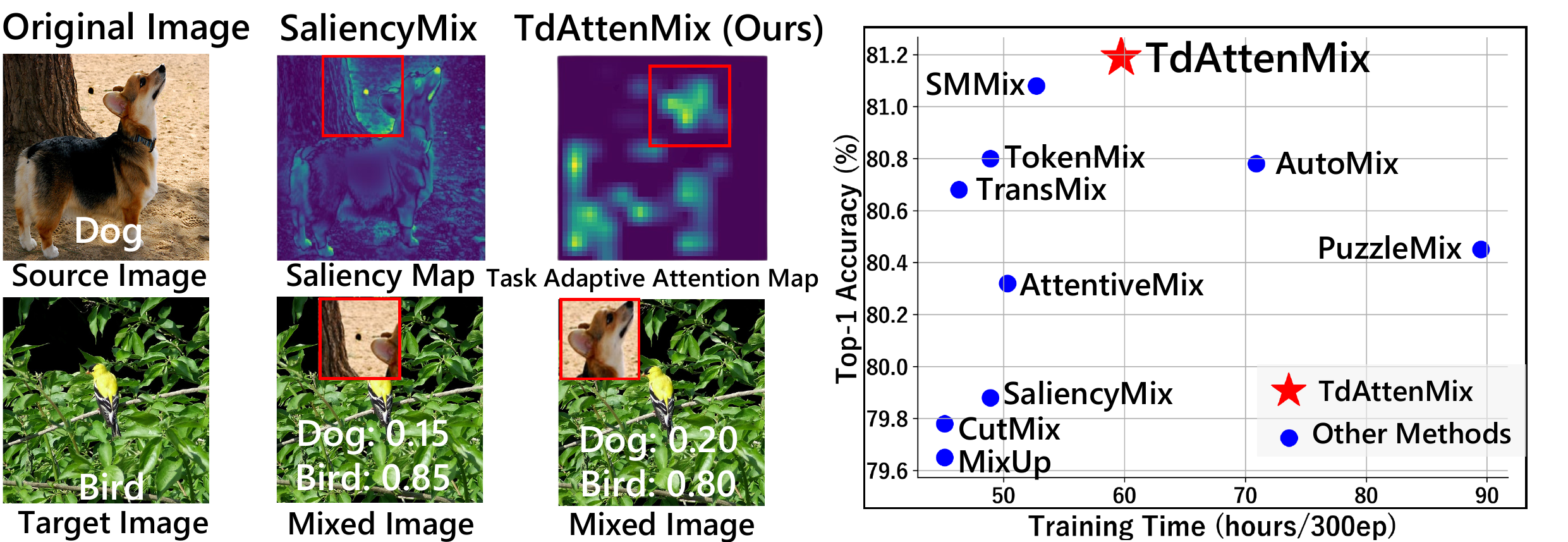}
    \vspace{-0.5em}
    \caption{\textbf{Left:} SaliencyMix \emph{vs.} TdAttenMix. Since SaliencyMix selects to crop the patch with the most salient region, it is distracted by 
    irrelevant dark stone. Our TdAttenMix balances top-down and bottom-up attention and thus picks salient areas consistent with the dog label. \textbf{Right:} Traing time \emph{vs.} accuracy with Deit-S on ImageNet-1k. TdAttenMix improves performance without the heavy computational overhead.}
    \label{fig:mix_example}
\end{figure}

\begin{figure*}[!t]
\centering
\includegraphics[width=1\textwidth]{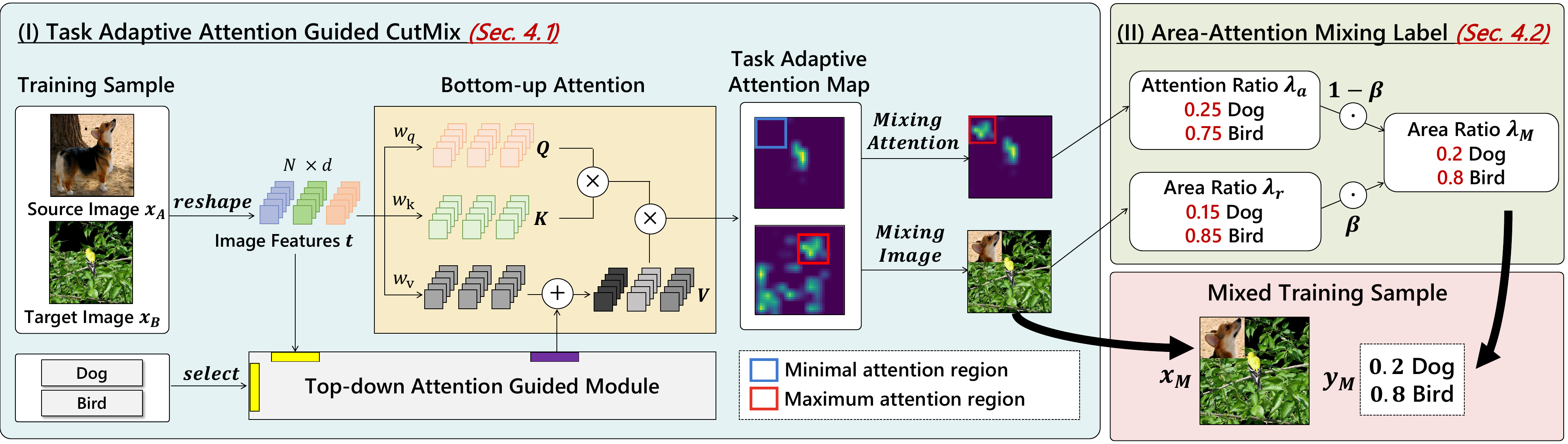}
\vspace{-0.5em}
\caption{The framework of TdAttenMix. (1)Task Adaptive Attention Guided CutMix: compute the task adaptive attention map via manipulating the bottom-up attention using our proposed Top-down Attention Guided Module and then uses the task adaptive attention map to crop the patch. (2)Area-Attention Label Mixing: adjust label mixing based on the ratio of attention and area.}

\label{fig:framework}
\vspace{-1.0em}
\end{figure*}

Since the randomness in CutMix~\cite{cutmix} ignores the spatial saliency, a group of saliency-based variants ~\cite{saliencymix, puzzlemix, attentivecutmix, automix, supermix, smmix} leverage the bottom-up attention as a supervisory signal. Bottom-up attention operates on raw sensory input and orients attention towards visual features of potential importance to calculate the saliency. This process only discovers \textit{what is where} in the world~\cite{schwinn2022behind}, which equally looks for all salient regions in the raw sensory input. Therefore, existing saliency-based variants based on bottom-up attention are easily distracted by high saliency regions that are, in fact, irrelevant to the target label. For instance,  the source image of Figure~\ref{fig:mix_example}  is dog, but SaliencyMix~\cite{saliencymix} become distracted by the dark rock  and crops the background, including only part of the dos's ear. 

Human vision entails more than just the determination of \textit{what is where}; it involves the development of internal representations that facilitate future actions. For instance, psychological research~\cite{buswell1935people, yarbus2013eye, belardinelli2015goal} found that human gaze, initially guided by bottom-up features, can be strongly influenced by the task at hand. Consequently, recent research proposes top-down mechanisms~\cite{absvit,schwinn2022behind} showing effectiveness in modeling human gaze patterns such as scanpaths~\cite{schwinn2022behind} and in enhancing downstream recognition tasks like classification and Vision Question Answering (VQA)~\cite{absvit}. For data mixing techniques, the labels of original data can be used naturally as the task at hand and the execution logic of human gaze can be modeled by the bottom-up features from the original image in conjunction with the high-level guidance of the original label. 

In this paper, inspired by the task guided mechanism of human gaze, we extend the saliency-based CutMix to a general framework that balances top-down and bottom-up attention to cut and mix the training samples. Bottom-up attention learns features from the original input and looks for all salient regions while top-down attention uses characteristics of the category as the current task to adjust attention. As illustrated in Figure~\ref{fig:mix_example}, our top-down attention mixup (TdAttenMix) crops the exemplary head region of the dog image and pastes it on the background area of the target bird image and finally obtains an image-label consistent mixed data that differs significantly from mixed data generated by SaliencyMix.

As portrayed in Figure~\ref{fig:framework}, our TdAttenMix involoves two steps: Task Adaptive Attention Guided CutMix and Area-Attention Label Mixing. The first step generalizes the bottom-up attention of saliency-based CutMix to the task adaptive attention via our proposed Top-down Attention Guided Module. When mixing the image, we use Max-Min Attention Region Mixing~\cite{smmix} to select maximum attention region from a source image and paste it onto the region with the minimal attention score in a target image. 

The second step determines the label mixing with Area-Attention Label Mixing module. Unlike the conventional approach of area-based label assignment, this module incorporates the area ratio of the mixed image along with the attention ratio of the task adaptive attention map. In the end, our TdAttenMix framework produces the mixed training sample  $(x_M, y_M)$ in Figure~\ref{fig:framework}. 

The saliency-based CutMix variants aim to produce a sufficient amount of image-label consistent mixed data. As far as current knowledge allows, existing methods lack a quantitative approach to assess image-label inconsistency. The core of quantitative analysis lies in establishing the correct ground-truth for label assignment. Motivated by the notion that gaze mirrors human vision~\cite{huang2020mutual},  we propose to use gaze attention on the ARISTO dataset~\cite{chrf} which collects the real gaze of participants when performing fine-grained recognition tasks. This data will be used to create mixed labels, serving as the ground truth to investigate the issue of image-label inconsistency.

The contribution of this paper is three-fold: 
\begin{itemize}[topsep=5pt]
\item[$\bullet$] By our knowledge, this paper for the first time proposes a Top-down Attention Guided Module to integrate human gaze for an artificial attention that balances both top-down and bottom-up attention to crop the task-relevant patch and adjust the label mixing ratio.
\item[$\bullet$] Extensive experiments demonstrate the TdAttenMix boosts the performance and achieve state-of-the-art top-1 accuracy in CIFAR100, Tiny-ImageNet, CUB-200 and ImageNet-1k. Moreover, as shown in Figure~\ref{fig:mix_example}, our TdAttenMix can achieve state-of-the-art top-1 accuracy without the heavy computational overhead.
\item[$\bullet$] We quantitatively explore the image-label inconsistency problem in image mixing. The proposed method effectively reduces the image-label inconsistency and improves the performance.
\end{itemize}

\section{Related Work}
\subsection{CutMix and its variants} \label{sec:cutmixvariants}
CutMix~\cite{cutmix} randomly crops a patch from the source image and pastes it onto the corresponding location in the target image, with labels being a linear mixture of the source and target image labels proportionate to the area ratio. Since random cropping ignores the regional saliency information, researchers leverage a series of saliency-based variants based on bottom-up attention. AttentiveMix~\cite{attentivecutmix} and SaliencyMix~\cite{saliencymix} guide mixing patches by saliency regions in the image (based on class activation mapping or a saliency detector\cite{montabone}). Subsequently, PuzzleMix~\cite{puzzlemix} and Co-Mixup~\cite{comixup} propose combinatorial optimization strategies to find optimal mixup that maximizes the saliency information. Then AutoMix~\cite{automix} adaptively generates mixed samples based on mixing ratios and feature maps in an end-to-end manner. Inspired by the success of Vision Transformer (ViT)\cite{vit}, TokenMixup~\cite{choi2022tokenmixup} is proposed to adaptive generate mixed images based on attention map. Moreover, concerning label assignment, recent studies have also adjusted label assignment by bottom-up attention. TransMix~\cite{transmix} mixes labels based on the class attention score and TokenMix~\cite{tokenmix} assigns content-based mixes labels on mixed images. Recently, SMMix~\cite{smmix} motivates both image and label enhancement by the bottom-up self-attention of ViT-based model under training itself. However, these existing variants, focusing either on enhancing saliency or adjusting label assignments, are reliant on bottom-up attention, which is susceptible to being distracted by salient but label-inconsistent background areas. To relieve label inconsistency, we introduce task adaptive top-down attention into CutMix variants for the first time and propose our framework TdAttenMix.

\subsection{Computational modeling of Attention}
Computational modeling of human visual attention intersects various disciplines such as neuroscience, cognitive psychology, and computer vision. Biologically-inspired attention mechanisms can enhance the interpretability of artificial intelligence~\cite{vuyyuru2020biologically}. The attention can be categorized into bottom-up and top-down mechanisms~\cite{connor2004visual}. Initially, the focus was primarily on computational modeling of bottom-up attention. Based on the Treisman's seminal work describing the feature integration theory~\cite{treisman1980feature}, current approaches assume a central role for the saliency map. Within the theory, attention shifts are generated from the saliency map using the winner-take-all algorithm~\cite{koch1985shifts}. Consequently, the majority of studies have focused on improving the estimation of the saliency map~\cite{borji2012state, riche2013saliency}. Recently self-attention~\cite{vit} is a stimulus-driven approach that highlights all the salient objects in an image, representing a typical bottom-up attention mechanism. With the advent of increasingly large eye-tracking datasets~\cite{chrf, jiang2015salicon}, researchers have been inspired to explore task-guided top-down attention. Shi et al.~\cite{absvit} propose a top-down modulated ViT model by mimicking the task-guided mechanism of human gaze. Shiwinn et al.~\cite{schwinn2022behind} impose a biologically-inspired foveated vision constraint to neural networks to generate human-like scanpaths without training for this object. As for CutMix variants, previous saliency-based methods have utilized bottom-up attention to optimize cropping regions, whereas we explore the use of task-adaptive top-down attention to obtain a cropped region that is more consistent with the label.

\section{Preliminary}
\textbf{CutMix augmentation.} CutMix~\cite{cutmix} is a simple data augmentation technique that combines two pairs of input and labels. $x$ and $y$ represent a training image and its corresponding label, where $x \in \mathbb{R}^{H \times W \times C}$. To create a new augmented training sample $(x_M, y_M)$, CutMix~\cite{cutmix} utilizes a source image-label pair $(x_{A}, y_{A})$ and a target image-label pair $(x_{B}, y_{B})$. Mathematically, this can be expressed as follows:
\begin{equation} \label{eq1}
  x_M = M \odot x_{A} + (1 - M) \odot x_{B}
\end{equation}
\begin{equation} \label{eq:arealabel}
  y_M = \lambda_r y_{A} + (1 - \lambda_r) y_{B}
\end{equation}

$M \in \{0,1\}^{H \times W}$ denotes a rectangular binary mask that indicates where to drop or keep in the two images, $\odot$ is element-wise multiplication, and $\lambda_r$ indicates the area ratio of $x_A$ in mixed image $x_M$, \textit{i.e.}, $\lambda_r = \frac{\sum M}{HW}$.

\section{Framework of TdAttenMix}
This section formally introduces our TdAttenMix, a general image mixing framework that balances top-down and bottom-up attention to simulate the task-guided mechanism of human gaze to crop the patch and adjust the label mixing ratio. Figure~\ref{fig:framework} illustrates an overview of our proposed TdAttenMix. Details are given below.

\subsection{Task Adaptive Attention Guided CutMix}
We want to simulate the execution logic of human gaze, which is initially guided by bottom-up features and then strongly influenced by the current task.

\textbf{Bottom-up Attention.} We divide the source image $x_A$ and the target image $x_B$ into non-overlapping patches of size $P \times P$. Each image yields a total of $N = \frac{H}{P} \times \frac{W}{P}$ patches. Consequently, $x_A$ and $x_B$ are restructured as $t_A, t_B \in \mathbb{R}^{N \times (P^2C)}$, where each row corresponds to a token and $d = P^2C$. As illustrated in Figure~\ref{fig:framework}, we follow SMMix~\cite{smmix} which obtains the attention map across all the image tokens for the bottom-up attention~\cite{vit}. We obtain $Q = tw_q$, $K = tw_k$, and $V = tw_v$, where $w_q \in \mathbb{R}^{d \times d}$, $w_k \in \mathbb{R}^{d \times d}$, and $w_v \in \mathbb{R}^{d \times d}$ represent the learnable parameters of the fully-connected layers.

\begin{figure}
    \centering
    \includegraphics[width=1\linewidth,height=0.45\linewidth]             {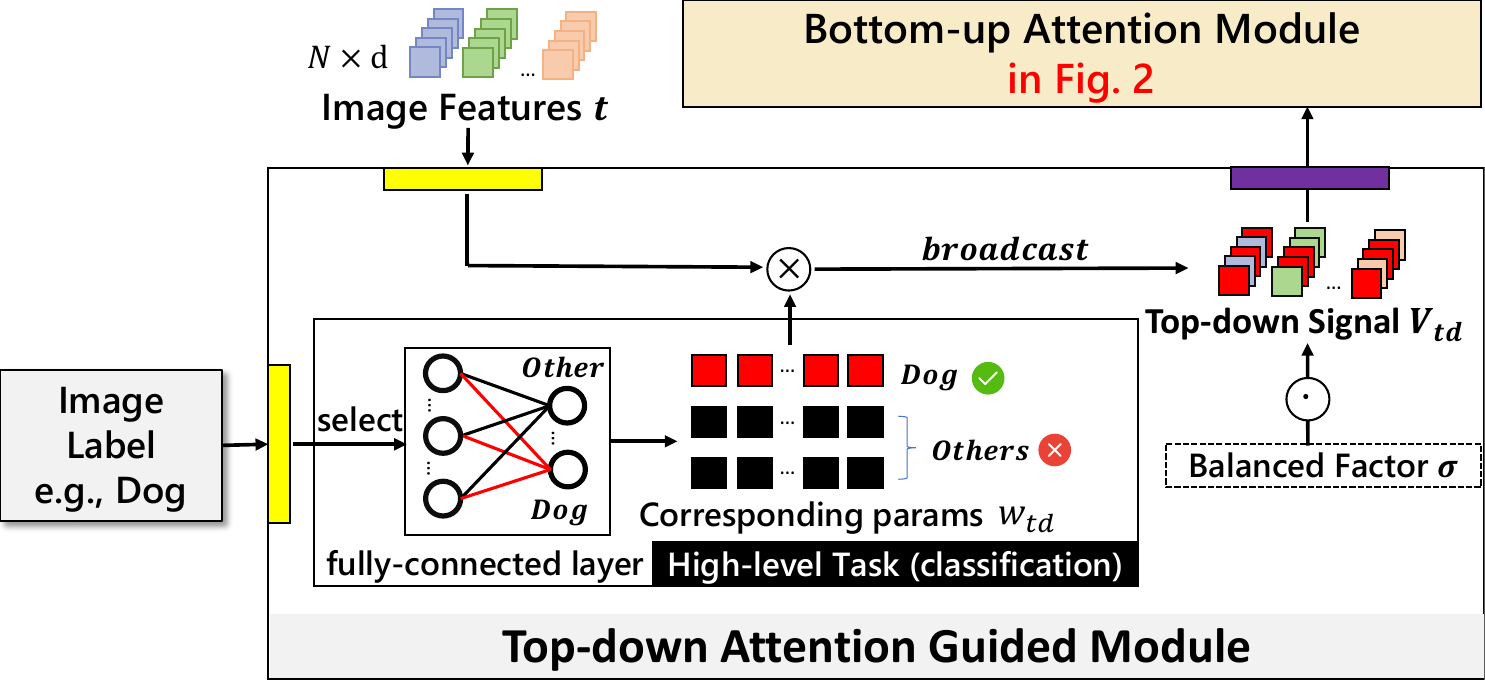}
    \vspace{-0.5em}
    \caption{To simulate the top-down mechanism, we designed the Top-down Attention Guided Module by using the image label as the high-level task information to guide image feature generation, resulting in what we refer to as the "top-down signal." This top-down signal then constrains bottom-up attention to focus on regions related to the image label.}
    \label{fig:top_down_module}
\end{figure}

\textbf{Top-down Attention Guided Module.} The Top-down Attention Guided Module we propose is depicted in Figure~\ref{fig:top_down_module}. The current task at hand is the classification task. Then we extract the corresponding parameters $w_{td} \in \mathbb{R}^{d \times 1}$ from the final fully-connected layer of Vision Transformer, which is based on the current label. The parameter matrix from this layer mirrors the relationship between feature and category mapping. Thus, we can acquire the high-level guidance $V_{td}$ tied to a specific category by calculating it with the image feature $t$. The theory that top-down attention can be implemented by simply augmenting $V_{td}$ to $V$ with $K$ and $Q$ remaining constant was introduced by Shi et al. \cite{absvit}. We ensure that the dimensionality of $V_{td}$ and $V$ is consistent through broadcasting. Furthermore, we accommodated a tunable parameter called balanced factor $\sigma$ within our framework to manage the top-down features $V_{td}$. If $\sigma = 0$, our attention map correlates with the preceding bottom-up attention utilized by SMMix~\cite{smmix}. If $\sigma = 1$, the attention map is finalized by integrating the bottom-up features with the top-down features. As a result, we calculate the task adaptive balanced attention as follows:
\begin{equation} \label{eq:v_td}
    V_{td} = \sigma \times broadcast(tw_{td}) 
\end{equation}
\begin{equation}
    V = V + V_{td}
\end{equation}
\begin{equation} \label{eq3}
   Attention(Q, K, V) = Softmax(\frac{QK^T}{\sqrt{d}}) V
\end{equation}

Subsequently, the resulting task adaptive attention maps, $\alpha_A \in \mathbb{R}^{\frac{H}{P} \times \frac{W}{P}}$ and $\alpha_B \in \mathbb{R}^{\frac{H}{P} \times \frac{W}{P}}$ corresponding to $x_A$ and $x_B$, are obtained after reshaping operation. Now our attention map is task adaptive which focuses on the object indicated by the current task while ignoring irrelevant high saliency objects. The criterion for cropping is determined by the sum of attention scores within a given region. Then we identify the region with maximum attention scores in the source image, and the region with the minimal attention sources in the target image. Specifically, the center indices are defined as:
\begin{equation}
    i_s, j_s = \underset{i, j}{argmax} \underset{p, q}{\sum} \alpha_A^{i+p-\lfloor \frac{h}{2} \rfloor, j + p-\lfloor \frac{w}{2} \rfloor}
\end{equation}
\begin{equation}
    i_t, j_t = \underset{i, j}{argmin} \underset{p, q}{\sum} \alpha_B^{i+p-\lfloor \frac{h}{2} \rfloor, j + p-\lfloor \frac{w}{2} \rfloor}
\end{equation}

$h = \lfloor \delta \frac{H}{P} \rfloor$, $w = \lfloor \delta  \frac{W}{P}\rfloor$, $ \delta = Uniform(0.25, 0.75)$, $p \in \{0, 1, ..., h-1\}$, and $q \in \{0, 1, ..., w-1\}$. Then we use Max-Min Attention Region Mixing~\cite{smmix} which uses the maximum attention region to replace the minimal attention region to obtain the new mixed training image $x_M$ as follows:
\begin{equation}
    x_M = x_B 
\end{equation}
\begin{equation}
    x_M^{i_t + p - \lfloor \frac{h}{2} \rfloor, j_t + q - \lfloor \frac{w}{2} \rfloor} = x_A^{i_s + p - \lfloor \frac{h}{2} \rfloor, j_s + q - \lfloor \frac{w}{2} \rfloor}
\end{equation}

\begin{figure}
    \centering
    \includegraphics[width=0.95\linewidth,height=0.4\linewidth]             {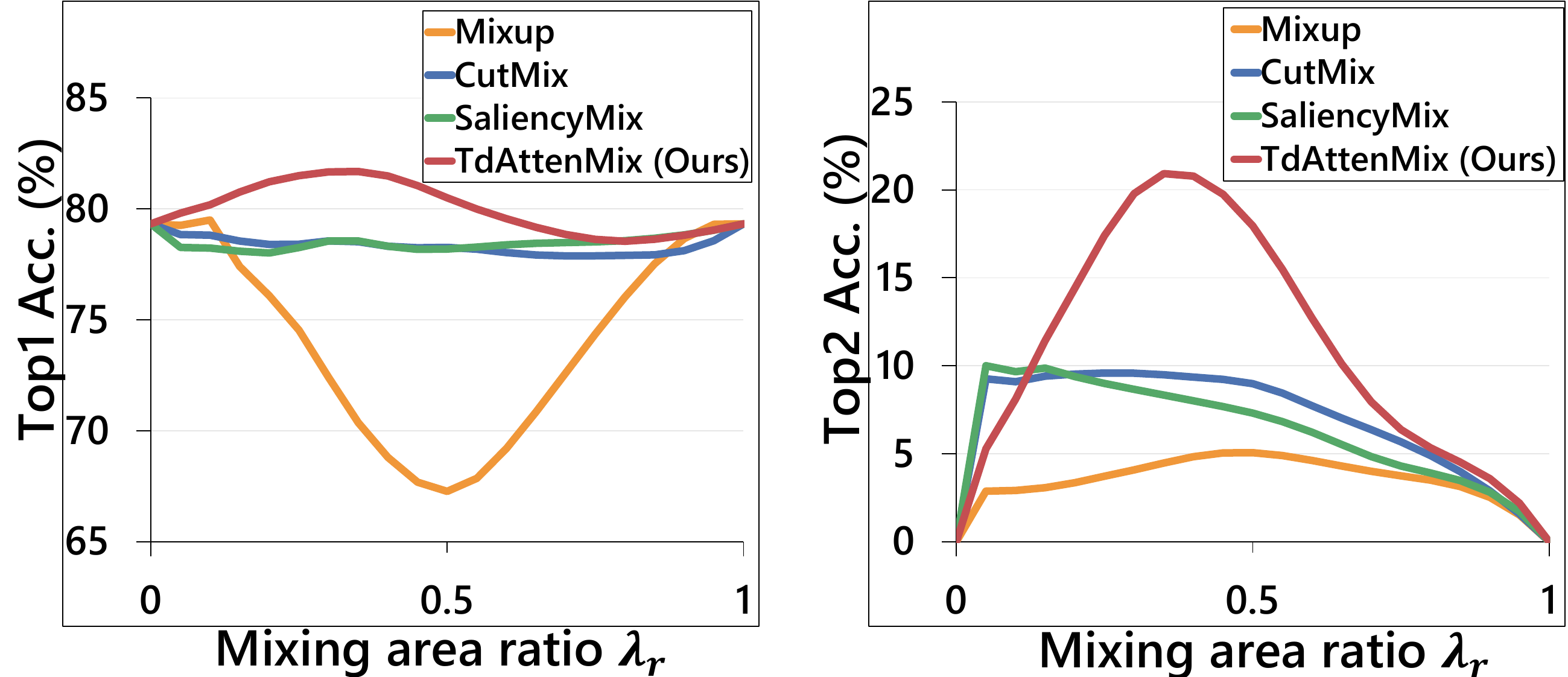}
    \vspace{-0.5em}
    \caption{Top-1 accuracy of mixed data. Prediction is counted as correct if the top-1 prediction belongs to $\{y_A, y_B\}$; Top-2 accuracy is calculated by counting the top-2 predictions are equal to $\{y_A, y_B\}$. $\lambda_r$ indicates the area ratio of $x_A$ in mixed image $x_M$.}
    \label{fig:top1top2}
\end{figure}

To verify the validity of the obtained mixed image $x_M$, we examined the prediction accuracy on the mixed image $x_M$. As graphically represented in Figure~\ref{fig:top1top2}, the prediction accuracy of mixed samples can be significantly improved by our method. Notably, for the top-2 accuracy, our TdAttenMix achieves 20.92\% while SaliencyMix~\cite{saliencymix} only reaches 10.00\%. This demonstrates that we obtain a mixed image consistent with the labels of source and target images.

\subsection{Area-Attention Label Mixing}
To enhance the precision of the mixed label $y_M$, based on the area ratio (Eq.~\ref{eq:arealabel}) used by CutMix~\cite{cutmix}, we adjust the area ratio using the attention scores of $\alpha_A$ and $\alpha_B$ at their respective positions within the mixed image $x_M$. More specifically, the final mixing ratio $\lambda$ are defined as follows:
\begin{equation}
    \lambda_r = \frac{hwP^2}{HW}
\end{equation}
\begin{equation}
        Att_A = \underset{p, q}{\sum} \alpha_A^{i_s + p - \lfloor \frac{h}{2} \rfloor, j_s + q - \lfloor \frac{w}{2} \rfloor} 
\end{equation}
\begin{equation}
        Att_B = \sum \alpha_B - \underset{p, q}{\sum} \alpha_B^{i_t + p - \lfloor \frac{h}{2} \rfloor, j_t + q - \lfloor \frac{w}{2} \rfloor}
\end{equation}
\begin{equation}
        \lambda_a = \frac{Att_A}{Att_A + Att_B}
\end{equation}
\begin{equation}
    \lambda = \beta \lambda_r + (1 - \beta) \lambda_a
\end{equation}

$\lambda_r$ is the area ratio of $x_A$ in mixed image $x_M$, $Att_A$ and $Att_B$ is the sum of the task adaptive attention scores at the positions corresponding to $x_M$ in $\alpha_A$ and $\alpha_B$, $\lambda_a$ is the attention ratio of $x_A$ in mixed image $x_M$, $\beta$ = 0.5, $\lambda$ is the final mixing ratio of $x_A$ in the mixed image $x_M$. The final mixed label $y_M$ is then defined as follows:
\begin{equation} \label{eq:mixlabel}
    y_M =  \lambda y_{A} + (1-\lambda) y_{B}
\end{equation}

Then we obtain the new mixed training sample $(x_M, y_M)$.

\subsection{Training Objective}
Our TdAttenMix framework is independent on any training model and can be used on various mainstream structures. When deployed on a ResNet-based architecture, we employ the standard classification loss and the consistency constraint losses proposed in SMMix~\cite{smmix}. The traditional classification loss is defined as follows, where $Y_M$ is the prediction distribution of mixed images. :
\begin{equation}
    L_{cls} = CE(Y_M, y_M)
\end{equation}

Then we require feature consistency constraint losses~\cite{smmix}, which help features of the mixed images fall into a consistent space with those of the original unmixed images. The feature consistency constraint losses in our TdAttenMix is:
\begin{equation}
    L_{con} = L_1(Y_M, \lambda Y_A + (1-\lambda)Y_B)
\end{equation}

$Y_A$ and $Y_B$ is the prediction distributions  of unmixed images $x_A$ and $x_B$. Overall, the training loss is then written as follows: 
\begin{equation}
    L_{total} = L_{cls} + L_{con}
\end{equation}

When deployed on a ViT-based architecture, we use the same loss function like SMMix~\cite{smmix}, which proves to be effective in learning features for mixed samples:
\begin{equation}
    L_{fine} = \frac{1}{2}(CE(Y_A, y_A) + CE(Y_B, y_B))
\end{equation}
\begin{equation}
    L_{total} = L_{cls} + L_{fine} + L_{con}
\end{equation}

\section{Experiments}
We evaluate TdAttenMix in four aspects: 1) Evaluating image classification tasks on eight different benchmarks, 2) transferring pre-trained models to two downstream tasks, 3) Evaluating the robustness on three scenarios including occlusion and two out-of-distribution datasets. (4) In addition, we have conducted the first quantitative study on the effectiveness of saliency-based methods in reducing image-label inconsistency. Our TdAttenMix is highlighted in gray, and \textbf{bold} denotes the best results.

\begin{table*}[!t]
    \small
    \centering

    \begin{tabular}{ccccccc|cc}
    \hline
    Dataset & \multicolumn{2}{c}{CIFAR100} & \multicolumn{2}{c}{Tiny-ImageNet} & \multicolumn{2}{c}{CUB-200} & \multicolumn{2}{|c}{ImageNet-1k} \\
    Network                                & R-18    &  RX-50  & R-18 &  RX-50      & R-18 &  RX-50   & R-18 &  Deit-S                    \\
    \hline
     Vanilla~\cite{li2023openmixup}        & 78.04   & 81.09   & 61.68   & 65.04    & 77.68  & 83.01      & 70.04   & 75.70                             \\
     SaliencyMix~\cite{saliencymix}        & 79.12   & 81.53   & 64.60   & 66.55    & 77.95    & 83.29      & 69.16   & 79.88                         \\
     MixUp~\cite{mixup}                    & 79.12   & 82.10   & 63.86   & 66.36    & 78.39  & 84.58      & 69.98  & 79.65                        \\
     CutMix~\cite{cutmix}                  & 78.17   & 81.67   & 65.53   & 66.47    & 78.40    & 85.68     & 68.95   & 79.78                     \\
     MainfoldMix~\cite{mainfoldmix}        & 80.35   & 82.88   & 64.15   & 67.30    & 79.76   & 86.38      & 69.98   & -                    \\
     SmoothMix~\cite{smoothmix}            & 78.69   & 80.68   & 66.65   & 69.65     & -   &   -        & -   & -                      \\
     AttentiveMix~\cite{attentivecutmix}   & 78.91   & 81.69   & 64.85   & 67.42    & -       &   -          & 68.57   & 80.32                   \\
     PuzzleMix~\cite{puzzlemix}            & 81.13   & 82.85   & 65.81   & 67.83    & 78.63   & 84.51     & 70.12   & 80.45             \\
     Co-Mixup~\cite{comixup}               & 81.17   & 82.91   & 65.92   & -        & -      &   -     & -   & -                          \\
     GridMix~\cite{gridmix}                & 78.72   & 81.11   & 65.14   & 66.53    & -     &   -    & -   & -                         \\
     TransMix~\cite{transmix}              &  -      & -       & -       & -        & -       & -      & -    & 80.68         \\
     TokenMix~\cite{tokenmix}              &  -      & -       & -       & -        & -       & -      & -    & 80.80         \\
     SMMix~\cite{smmix}                    &  -      & -       & -       & -        & -       & -      & -    & 81.08         \\
     AutoMix~\cite{automix}                & 82.04   & 83.64   & 67.33   & 70.72    & 79.87   & 86.56  & 70.50     & 80.78         \\
     \rowcolor[gray]{0.9} TdAttenMix (Ours)  & \textbf{82.36}  & \textbf{84.03}      & \textbf{67.47}    & \textbf{70.80} 
                                             & \textbf{80.71}  & \textbf{86.72}  & \textbf{70.74}    & \textbf{81.19}   \\ 
     \hline
     Gain                                  & \color[rgb]{0,0.39,0}{+3.24}    & \color[rgb]{0,0.39,0}{+2.50}   
                                           & \color[rgb]{0,0.39,0}{+2.87}  & \color[rgb]{0,0.39,0}{+4.25}    
                                           & \color[rgb]{0,0.39,0}{+2.76}   & \color[rgb]{0,0.39,0}{+3.43} 
                                           & \color[rgb]{0,0.39,0}{+1.58}  & \color[rgb]{0,0.39,0}{+1.31} \\
    \hline
    \end{tabular}
    \caption{Image classification top-1 accuracy (\%) on CIFAR-100, Tiny-ImageNet, CUB-200 and ImageNet-1k. We get the performance of previous methods from the OpenMixup~\cite{li2023openmixup} benchmark. Gain: indicates the performance improvement compared with SaliencyMix.}
    \label{tab:smallscale_imagenet}
    \vspace{-1.5em}
\end{table*}

\subsection{Small-scale Classification}\label{sec:smallscaleresult}

In small-scale classification we use ResNet-18~\cite{resnet} and ResNext-50~\cite{resnext} to compare the performance. Hyperparameter settings are in the section 1 of the Supplementary. Table~\ref{tab:smallscale_imagenet} shows small-scale classification results on CIFAR-100, Tiny-ImageNet and CUB-200. Compared to the previous SOTA methods, TdAttenMix consistently surpasses AutoMix (+0.08$\sim$ +0.84), PuzzleMix (+1.18 $\sim$ +2.97), MainfoldMix (+0.34 $\sim$ +3.35) based on various ResNet architectures. Moreover, TdAttenMix noticeably exhibits a significant gap with SaliencyMix (+2.50 $\sim$ + 4.25).

\subsection{ImageNet Classification} 

Table~\ref{tab:smallscale_imagenet} validates the  performance advantage of TdAttenMix over other methods. In particular, TdAttenMix boosts the top-1 accuracy by more than +1\% in ResNet-18~\cite{resnet} and Deit-S~\cite{deit} compared with the SaliencyMix baseline and achieves the sota result. It can be noted that TransMix, TokenMix and SMMix also exhibit good top-1 accuracy, but they are limited to ViT-special methods, causing them incompatible with all mainstream architectures (e.g., ResNet). We provide more comparisons with ViT-special methods in Section 2 of the Supplementary, and additional experiments have proven the effectiveness of our TdAttenMix. On the contrary, our TdAttenMix is an independent data argumentation method which is compatible with mainstream architectures.

\begin{table}[!t]
    \small
    \centering
    \setlength\tabcolsep{1pt}
    \begin{tabular}{cccc}
    \hline
         & \multicolumn{2}{c}{Semantic segmentation} & \multicolumn{1}{c}{WSAS} \\
     Models                                 & mIoU (\%) & mAcc (\%) & Segmentation JI (\%)    \\
     \hline
     Deit-S                                 &  31.6      &  44.4       &  29.2                                \\
     \rowcolor[gray]{0.9} TdAttenMix-Deit-S      &  \textbf{33.2}    & \textbf{46.9}  & \textbf{32.5}   \\
    \hline
     Gain   & \color[rgb]{0,0.39,0}{+1.6}    & \color[rgb]{0,0.39,0}{+2.5}   & \color[rgb]{0,0.39,0}{+3.3}\\
    \hline
    \end{tabular}
    \caption{Downstream tasks. Transferring the pre-trained models to semantic segmentation task using UperNet with DeiT backbone on ADE20k dataset; Segmentation JI denotes the jaccard index for weakly supervised automatic segmentation (WSAS) on Pascal VOC.}
    \label{tab:downstream}
\end{table}

\subsection{Downstream Tasks}

\textbf{Semantic segmentation.} We use ADE20k~\cite{zhou2017scene} to evaluate the performance of semantic segmentation task. ADE20k is a challenging scene parsing dataset covering 150 semantic categories, with 20k, 2k, and 3k images for training, validation and testing. We evaluate DeiT backbones with UperNet~\cite{upernet}. As shown in Table~\ref{tab:downstream}, TdAttenMix improves Deit-S for +1.6\% mIoU and +2.5\% mAcc.

\textbf{Weakly supervised automatic segmentation (WSAS).} We compute the Jaccard similarity over the PASCAL-VOC12 benchmark~\cite{voc}. The attention masks generated from TdAttenMix-DeiT-S or vanilla DeiT-S are compared with ground-truth on the benchmark. The evaluated scores can quantitatively help us to understand if TdAttenMix has a positive effect on the quality of attention map. As shown in Table~\ref{tab:downstream}, TdAttenMix improves Deit-S for +3.3\%.

\begin{figure*}[!t]
\centering
\includegraphics[width=1\textwidth]{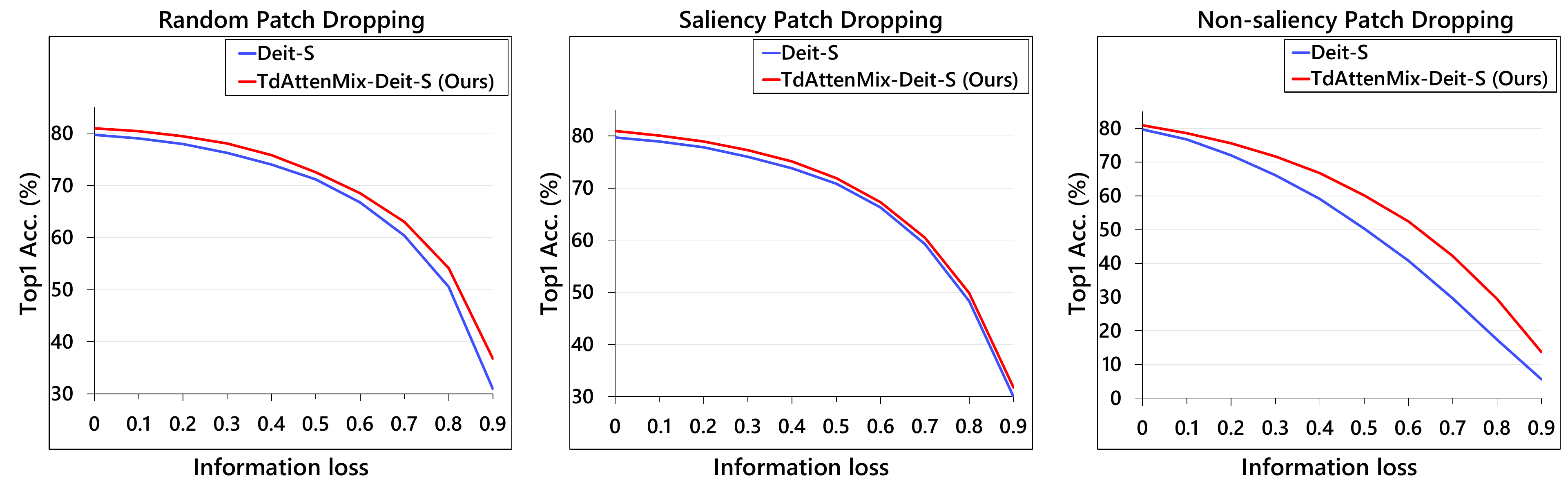}
\vspace{-0.5em}
\caption{Robustness against occlusion. Model  robustness against occlusion with different information loss ratios is studied. 3 patch dropping settings: Random Patch Dropping (left), Salient Patch Dropping (middle), and Non-Salient Patch Dropping (right) are considered.}
\label{fig:robustness_occlusion} 
\vspace{-1.0em}
\end{figure*}

\subsection{Robustness Analysis}

\textbf{Robustness to Occlusion.} Naseer et al.~\cite{naseer2021intriguing} studies whether ViTs perform robustly in occluded scenarios, where some of most of the image content is missing. Following~\cite{naseer2021intriguing}, we showcase the classification accuracy on ImageNet-1k validation set with three dropping settings. (1) Random Patch Dropping. (2) Salient (foreground) Patch Dropping. (3) Non-salient (background) Patch Dropping. As depicted in Figure~\ref{fig:robustness_occlusion}, Deit-S augmented with TdAttenMix outperforms the standard Deit-S across all occlusion levels.

\begin{table}[!t]
    \small
    \centering
    \setlength\tabcolsep{0pt}
    \begin{tabular}{ccccc}
    \hline
         & \multicolumn{3}{c}{Nat. Adversarial Example} & \multicolumn{1}{c}{Out-of-Dist} \\
     Models                                 & Top-1 Acc. & Calib-Error$\downarrow$ & AURRA  &   AUPR    \\
     \hline
     Deit-S                                 &  19.1      &  32.0       &  23.8   &  20.9                                 \\
     \rowcolor[gray]{0.9} TdAttenMix-Deit-S      &  \textbf{22.0}    & \textbf{30.4}  & \textbf{29.7}  & \textbf{22.0}   \\
    \hline
     Gain   & \color[rgb]{0,0.39,0}{+2.9}    & \color[rgb]{0,0.39,0}{+1.6}   & \color[rgb]{0,0.39,0}{+5.9}  & \color[rgb]{0,0.39,0}{+1.1}\\
    \hline
    \end{tabular}
    \caption{Model's robustness against natural adversarial examples on ImageNet-A and out-of-distribution examples on ImageNet-O.}
    \label{tab:imagenet-ao}
\end{table}

\textbf{Out-of-distribution Datasets.} We evaluate our TDAttenMix on two out-of-distribution datasets. (1) The ImageNet-A dataset~\cite{hendrycks2021natural}. The metric for assessing classifiers' robustness to adversarially filtered examples includes the top-1 accuracy, Calibration Error (CalibError)~\cite{hendrycks2021natural,kumar2019verified}, and Area Under the Response Rate Accuracy Curve (AURRA)~\cite{hendrycks2021natural}. (2) The ImageNet-O~\cite{hendrycks2021natural}. The metric is the area under the precision-recall curve (AUPR) \cite{hendrycks2021natural}. Table~\ref{tab:imagenet-ao} indicates that TdAttenMix can have consistent performance gains over vanilla Deit-S on the out-of-distribution data.

\begin{table}[!t]
    \small
    \centering
    \begin{tabular}{cc}
    \hline
     Method              &  Inconsistency$\downarrow$  \\
     \hline
     CutMix~\cite{cutmix}              &  26.2   \\
     SaliencyMix~\cite{saliencymix}         &  18.9   \\
     TdAttenMix-Bottom-up    &  19.0 \\
     \rowcolor[gray]{0.9} TdAttenMix      &  \textbf{18.4} \\
    \hline
    Gain    & \color[rgb]{0,0.39,0}{+7.8}  \\
    \hline
    \end{tabular}
    \caption{Image-label inconsistency of different saliency-based CutMix variants. TdAttenMix-Bottom-up represents the settings of $\sigma$ to 0 to control the task adaptive balanced attention of TdAttenMix as the standard bottom-up attention. Gain: reduction of error.}
    \label{tab:inconsist}
\end{table}

\subsection{Image-label Inconsistency Analysis}
Previous image mixing methods did not quantitatively validate the image-label inconsistency.Motivated by the fact that gaze reflects human vision~\cite{huang2020mutual}, we propose using the mixed label, which is based on gaze attention, as the ground-truth to validate the problem of image-label consistency. For our experiments, we utilize ARISTO dataset~\cite{chrf} and the corresponding raw images. Since $\lambda$ determines the mixed label, the image-label inconsistency can be represented by the difference between the $\lambda$ and ground truth $\lambda_{gt}$ obtained by gaze attention for the same mixed image. So we define the metrics as:
\begin{equation}
    Inconsistency = |\lambda_{gt} - \lambda|
\end{equation}

$\lambda_{gt}$ is calculated based on the real human gaze, $\lambda$ is calculated based on different CutMix variants. As shown in Table~\ref{tab:inconsist}, the inconsistency is effectively reduced for saliency-based methods. Our TdAttenMix are +7.8 higher than random based CutMix~\cite{cutmix}. The result of TdAttenMix-Bottom-up using only bottom-up attention is close to the results obtained by SaliencyMix~\cite{saliencymix}. This may be due to neither TdAttenMix-Bottom-up nor SaliencyMix has task adaptive ability, thus image-label inconsistency will be stronger than our TdAttenMix. These experimental findings strongly support the notion that saliency-based CutMix variants enhance training by mitigating image-label inconsistency, with top-down attention being more effective than bottom-up attention.

\begin{figure}
    \centering
    \includegraphics[width=0.95\linewidth,height=0.55\linewidth]             {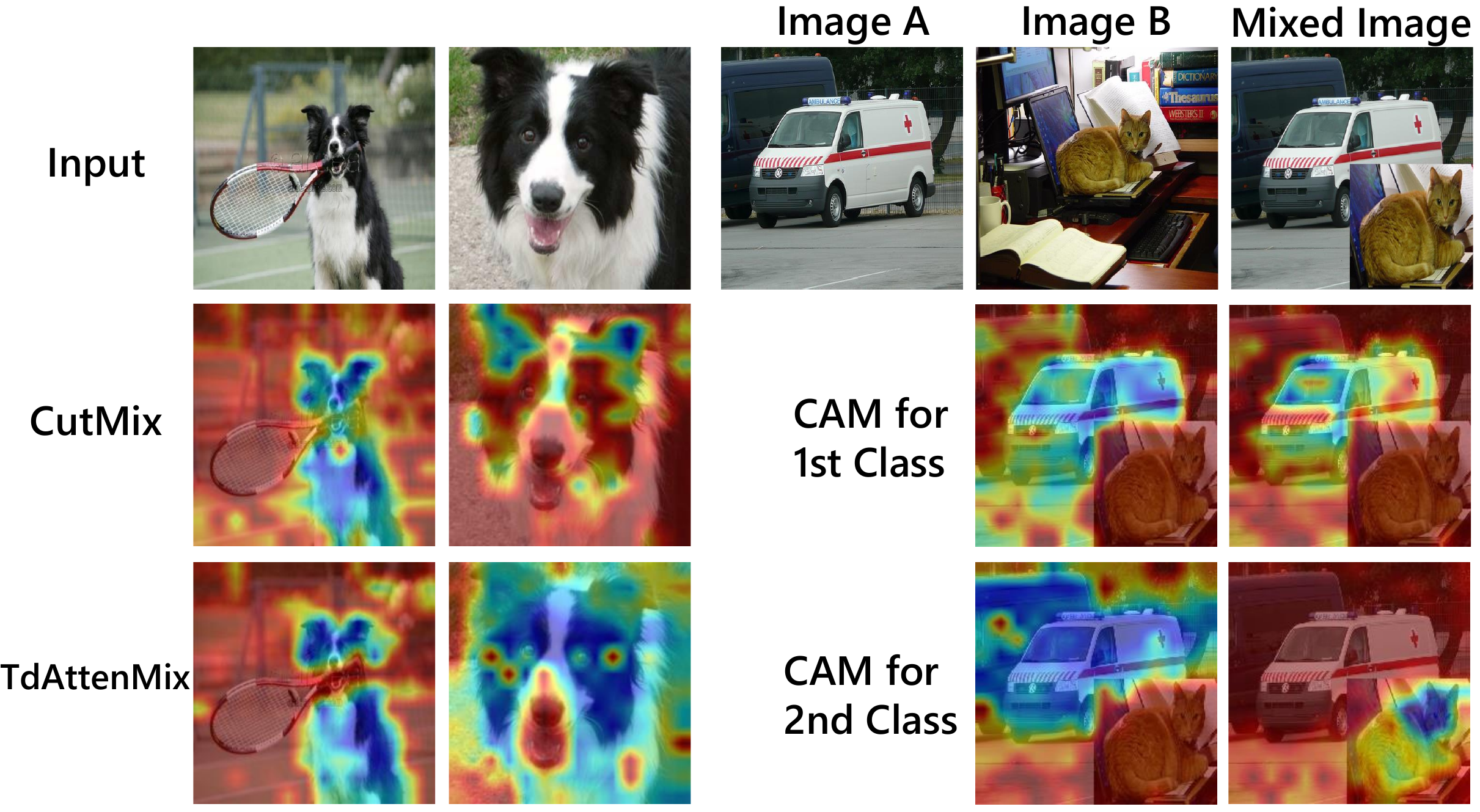}
    \vspace{-0.5em}
    \caption{We show the class activation map~\cite{cam} of the models trained with CutMix and TdAttenMix by testing on unmixed and mixed images, respectively. Left: locate objects in the unmixed images. Right: locate objects in the mixed images.}
    \label{fig:cam}
\end{figure}

\subsection{Visualization}
In Figure~\ref{fig:cam}, we visualize the class activation map~\cite{cam} of the models trained with CutMix and TdAttenMix. As shown in the left of Figure~\ref{fig:cam} that the TdAttenMix can locate object with more precision than the CutMix model in the unmixed images. Furthermore, the right of Figure~\ref{fig:cam} shows that for the mixed images, the TdAttenMix model can accurately locate objects from two different images. On the contrary, CutMix model focuses only on the class of image A. Our TdAttenMix is guided by task adaptive attention, which ensures that the information in the training data is sufficient enabling superior recognition capacity for mixed images. 

\subsection{Ablation Study} \label{ablation}
We conduct an ablation study to analyze our proposed TdAttenMix. We use ResNet-18~\cite{resnet} as the backbone and train it on CUB-200~\cite{cub}.

\begin{table}
    \small
    \centering
    \begin{tabular}{ccc}
    \hline
    Model & $\sigma$  &  Top-1 Acc.(\%) \\
    \hline
    \multirow{6}{*}{ResNet-18~\cite{resnet}} & 0 & 80.31 \\
    & 0.5 & 80.60 \\
    &\textbf{1} & \textbf{80.71} \\
    & 2 & 80.29 \\
    & 3 & 79.89 \\
    & 4 & 79.50 \\
    \hline
    \end{tabular}
    \caption{Control of task adaptive balanced attention. As shown in Eq.~\ref{eq:v_td}, the task adaptive balanced attention can be controlled by $\sigma$ which when $\sigma=0$ represents standard bottom-up attention.}
    \label{tab:bottomtotop}
\end{table}

\begin{table}
    \small
    \centering
    \begin{tabular}{ccc}
    \hline
    Model & $\beta$  &  Top-1 Acc.(\%) \\
    \hline
    \multirow{6}{*}{ResNet-18~\cite{resnet}} & 0 & 80.20 \\
    & 0.3 & 80.29 \\
    &\textbf{0.5} & \textbf{80.71}\\
    & 0.7 & 80.19 \\
    & 1 & 80.27 \\
    & random & 80.29 \\
    \hline
    \end{tabular}
    \caption{Mix ratio $\beta$ of area-attention label mixing.}
    \label{tab:beta}
\end{table}

\textbf{Control of Task Adaptive Balanced Attention.} Our TdAttenMix balances top-down and bottom-up attention by adjusting the top-down signal $V_{td}$, enabling a shift from standard bottom-up to top-down attention.We evaluate three different task adaptive balanced attention strategies: 1) $\sigma = 0$, 2) $\sigma = 0.5$, 3) $\sigma = 1$, 4) $\sigma = 2$, 5) $\sigma = 3$, 6) $\sigma = 4$. These strategies represent a gradual increase in task adaptive ability when bottom-up features are sufficient. This is consistent with the execution logic of human gaze, in which the top-down signal on top of the bottom-up features directs attention to achieve the best results.

\textbf{Mix ratio $\beta$ of area-attention label mixing.} $\beta$ determines the ratio of area-attention label mixing. We evaluate the performance for several values of $\beta$: 1) fixed as 0, which means only the area ratio is used to mix labels, 2) fixed as 0.3, 3) fixed as 0.5 which assigns equal weighting to the area ratio and attention ratio, 4) fixed as 0.7, 5) fixed as 1,  which means only the attention ratio is used to mix labels 6) random value of $\beta$, which means $\beta$ as a random number between 0 and 1. Table~\ref{tab:beta} shows that the best results are obtained when $\beta$ is set to 0.5.


\section{Conclusion}
This paper proposes TdAttenMix, a general and effective data augmentation framework. Motivated by the superiority of human gaze, we simulate the task-guided mechanism of human gaze to modulate attention. TdAttenMix introduces a new Top-down Attention Guided Module to balance bottom-up attention for task-related regions. Extensive experiments verify the effectiveness and robustness of TdAttenMix, which significantly improves the performance on various datasets and backbones. Furthermore, we quantitatively validate that our method and saliency-based methods can efficiently reduce image-label inconsistency for the first time.

\section*{Acknowledgements}
This work was supported by Beijing Natural Science Foundation (L242019). Dr. Lin Gu was also supported by JST Moonshot R\&D Grant Number JPMJMS2011 Japan.

\bibliography{aaai25}

\begin{thebibliography}{55}
\providecommand{\natexlab}[1]{#1}

\bibitem[{Baek, Bang, and Shim(2021)}]{gridmix}
Baek, K.; Bang, D.; and Shim, H. 2021.
\newblock GridMix: Strong regularization through local context mapping.
\newblock \emph{Pattern Recognition}, 109: 107594.

\bibitem[{Belardinelli, Herbort, and Butz(2015)}]{belardinelli2015goal}
Belardinelli, A.; Herbort, O.; and Butz, M.~V. 2015.
\newblock Goal-oriented gaze strategies afforded by object interaction.
\newblock \emph{Vision research}, 106: 47--57.

\bibitem[{Borji and Itti(2012)}]{borji2012state}
Borji, A.; and Itti, L. 2012.
\newblock State-of-the-art in visual attention modeling.
\newblock \emph{IEEE transactions on pattern analysis and machine intelligence}, 35(1): 185--207.

\bibitem[{Buswell(1935)}]{buswell1935people}
Buswell, G.~T. 1935.
\newblock How people look at pictures: a study of the psychology and perception in art.

\bibitem[{Chen, Fan, and Panda(2021)}]{Chen_2021_ICCV}
Chen, C.-F.~R.; Fan, Q.; and Panda, R. 2021.
\newblock CrossViT: Cross-Attention Multi-Scale Vision Transformer for Image Classification.
\newblock In \emph{Proceedings of the IEEE/CVF International Conference on Computer Vision (ICCV)}, 357--366.

\bibitem[{Chen et~al.(2022)Chen, Sun, He, Torr, Yuille, and Bai}]{transmix}
Chen, J.-N.; Sun, S.; He, J.; Torr, P.~H.; Yuille, A.; and Bai, S. 2022.
\newblock TransMix: Attend To Mix for Vision Transformers.
\newblock In \emph{Proceedings of the IEEE/CVF Conference on Computer Vision and Pattern Recognition (CVPR)}, 12135--12144.

\bibitem[{Chen et~al.(2023)Chen, Lin, Lin, Zhang, Chao, and Ji}]{smmix}
Chen, M.; Lin, M.; Lin, Z.; Zhang, Y.; Chao, F.; and Ji, R. 2023.
\newblock SMMix: Self-Motivated Image Mixing for Vision Transformers.
\newblock In \emph{Proceedings of the IEEE/CVF International Conference on Computer Vision (ICCV)}, 17260--17270.

\bibitem[{Cheng et~al.(2022)Cheng, Liang, Choi, Tao, Cao, Liu, and Zhang}]{Physicalattack}
Cheng, Z.; Liang, J.; Choi, H.; Tao, G.; Cao, Z.; Liu, D.; and Zhang, X. 2022.
\newblock Physical Attack on Monocular Depth Estimation with Optimal Adversarial Patches.
\newblock In Avidan, S.; Brostow, G.; Ciss{\'e}, M.; Farinella, G.~M.; and Hassner, T., eds., \emph{Computer Vision -- ECCV 2022}, 514--532. Cham: Springer Nature Switzerland.

\bibitem[{Choi, Choi, and Kim(2022)}]{choi2022tokenmixup}
Choi, H.~K.; Choi, J.; and Kim, H.~J. 2022.
\newblock Tokenmixup: Efficient attention-guided token-level data augmentation for transformers.
\newblock \emph{Advances in Neural Information Processing Systems}, 35: 14224--14235.

\bibitem[{Connor, Egeth, and Yantis(2004)}]{connor2004visual}
Connor, C.~E.; Egeth, H.~E.; and Yantis, S. 2004.
\newblock Visual attention: bottom-up versus top-down.
\newblock \emph{Current biology}, 14(19): R850--R852.

\bibitem[{Cui et~al.(2021)Cui, Yan, Cao, and Liu}]{Cui_2021_ICCV}
Cui, Y.; Yan, L.; Cao, Z.; and Liu, D. 2021.
\newblock TF-Blender: Temporal Feature Blender for Video Object Detection.
\newblock In \emph{Proceedings of the IEEE/CVF International Conference on Computer Vision (ICCV)}, 8138--8147.

\bibitem[{Cui et~al.(2022)Cui, Zhu, Gu, Qi, Li, Zhang, Zhang, and Harada}]{Cui23ECCV}
Cui, Z.; Zhu, Y.; Gu, L.; Qi, G.-J.; Li, X.; Zhang, R.; Zhang, Z.; and Harada, T. 2022.
\newblock Exploring Resolution and Degradation Clues as Self-supervised Signal for Low Quality Object Detection.
\newblock In Avidan, S.; Brostow, G.; Ciss{\'e}, M.; Farinella, G.~M.; and Hassner, T., eds., \emph{Computer Vision -- ECCV 2022}, 473--491. Cham: Springer Nature Switzerland.

\bibitem[{Dabouei et~al.(2021)Dabouei, Soleymani, Taherkhani, and Nasrabadi}]{supermix}
Dabouei, A.; Soleymani, S.; Taherkhani, F.; and Nasrabadi, N.~M. 2021.
\newblock SuperMix: Supervising the Mixing Data Augmentation.
\newblock In \emph{Proceedings of the IEEE/CVF Conference on Computer Vision and Pattern Recognition (CVPR)}, 13794--13803.

\bibitem[{Dosovitskiy et~al.(2021)Dosovitskiy, Beyer, Kolesnikov, Weissenborn, Zhai, Unterthiner, Dehghani, Minderer, Heigold, Gelly, Uszkoreit, and Houlsby}]{vit}
Dosovitskiy, A.; Beyer, L.; Kolesnikov, A.; Weissenborn, D.; Zhai, X.; Unterthiner, T.; Dehghani, M.; Minderer, M.; Heigold, G.; Gelly, S.; Uszkoreit, J.; and Houlsby, N. 2021.
\newblock An Image is Worth 16x16 Words: Transformers for Image Recognition at Scale.
\newblock In \emph{International Conference on Learning Representations}.

\bibitem[{Everingham et~al.(2015)Everingham, Eslami, Van~Gool, Williams, Winn, and Zisserman}]{voc}
Everingham, M.; Eslami, S.~A.; Van~Gool, L.; Williams, C.~K.; Winn, J.; and Zisserman, A. 2015.
\newblock The pascal visual object classes challenge: A retrospective.
\newblock \emph{International journal of computer vision}, 111: 98--136.

\bibitem[{He et~al.(2016)He, Zhang, Ren, and Sun}]{resnet}
He, K.; Zhang, X.; Ren, S.; and Sun, J. 2016.
\newblock Deep residual learning for image recognition.
\newblock In \emph{Proceedings of the IEEE conference on computer vision and pattern recognition}, 770--778.

\bibitem[{Hendrycks et~al.(2021)Hendrycks, Zhao, Basart, Steinhardt, and Song}]{hendrycks2021natural}
Hendrycks, D.; Zhao, K.; Basart, S.; Steinhardt, J.; and Song, D. 2021.
\newblock Natural adversarial examples.
\newblock In \emph{Proceedings of the IEEE/CVF Conference on Computer Vision and Pattern Recognition}, 15262--15271.

\bibitem[{Huang et~al.(2020)Huang, Cai, Li, Lu, and Sato}]{huang2020mutual}
Huang, Y.; Cai, M.; Li, Z.; Lu, F.; and Sato, Y. 2020.
\newblock Mutual context network for jointly estimating egocentric gaze and action.
\newblock \emph{IEEE Transactions on Image Processing}, 29: 7795--7806.

\bibitem[{Jiang et~al.(2015)Jiang, Huang, Duan, and Zhao}]{jiang2015salicon}
Jiang, M.; Huang, S.; Duan, J.; and Zhao, Q. 2015.
\newblock Salicon: Saliency in context.
\newblock In \emph{Proceedings of the IEEE conference on computer vision and pattern recognition}, 1072--1080.

\bibitem[{Kim et~al.(2020)Kim, Choo, Jeong, and Song}]{comixup}
Kim, J.; Choo, W.; Jeong, H.; and Song, H.~O. 2020.
\newblock Co-Mixup: Saliency Guided Joint Mixup with Supermodular Diversity.
\newblock In \emph{International Conference on Learning Representations}.

\bibitem[{Kim, Choo, and Song(2020)}]{puzzlemix}
Kim, J.-H.; Choo, W.; and Song, H.~O. 2020.
\newblock Puzzle mix: Exploiting saliency and local statistics for optimal mixup.
\newblock In \emph{International Conference on Machine Learning}, 5275--5285. PMLR.

\bibitem[{Koch and Ullman(1985)}]{koch1985shifts}
Koch, C.; and Ullman, S. 1985.
\newblock Shifts in selective visual attention: towards the underlying neural circuitry.
\newblock \emph{Human neurobiology}, 4(4): 219--227.

\bibitem[{Kumar, Liang, and Ma(2019)}]{kumar2019verified}
Kumar, A.; Liang, P.~S.; and Ma, T. 2019.
\newblock Verified uncertainty calibration.
\newblock \emph{Advances in Neural Information Processing Systems}, 32.

\bibitem[{Lee et~al.(2020)Lee, Zaheer, Astrid, and Lee}]{smoothmix}
Lee, J.-H.; Zaheer, M.~Z.; Astrid, M.; and Lee, S.-I. 2020.
\newblock Smoothmix: a simple yet effective data augmentation to train robust classifiers.
\newblock In \emph{Proceedings of the IEEE/CVF conference on computer vision and pattern recognition workshops}, 756--757.

\bibitem[{Li et~al.(2023)Li, Wang, Liu, Wu, Tan, Jin, and Li}]{li2023openmixup}
Li, S.; Wang, Z.; Liu, Z.; Wu, D.; Tan, C.; Jin, W.; and Li, S.~Z. 2023.
\newblock OpenMixup: A Comprehensive Mixup Benchmark for Visual Classification.
\newblock arXiv:2209.04851.

\bibitem[{Liu et~al.(2021{\natexlab{a}})Liu, Cui, Tan, and Chen}]{sg_net}
Liu, D.; Cui, Y.; Tan, W.; and Chen, Y. 2021{\natexlab{a}}.
\newblock SG-Net: Spatial Granularity Network for One-Stage Video Instance Segmentation.
\newblock In \emph{Proceedings of the IEEE/CVF Conference on Computer Vision and Pattern Recognition (CVPR)}, 9816--9825.

\bibitem[{Liu et~al.(2021{\natexlab{b}})Liu, Cui, Yan, Mousas, Yang, and Chen}]{liu2021densernet}
Liu, D.; Cui, Y.; Yan, L.; Mousas, C.; Yang, B.; and Chen, Y. 2021{\natexlab{b}}.
\newblock Densernet: Weakly supervised visual localization using multi-scale feature aggregation.
\newblock In \emph{Proceedings of the AAAI Conference on Artificial Intelligence}, volume~35, 6101--6109.

\bibitem[{Liu et~al.(2022{\natexlab{a}})Liu, Liu, Zhou, Li, and Liu}]{tokenmix}
Liu, J.; Liu, B.; Zhou, H.; Li, H.; and Liu, Y. 2022{\natexlab{a}}.
\newblock TokenMix: Rethinking Image Mixing for Data Augmentation in Vision Transformers.
\newblock In \emph{European Conference on Computer Vision}, 455--471.

\bibitem[{Liu et~al.(2022{\natexlab{b}})Liu, Zhou, Zhang, Bai, Gu, Yu, Zhou, and Hancock}]{chrf}
Liu, Y.; Zhou, L.; Zhang, P.; Bai, X.; Gu, L.; Yu, X.; Zhou, J.; and Hancock, E.~R. 2022{\natexlab{b}}.
\newblock Where to focus: Investigating hierarchical attention relationship for fine-grained visual classification.
\newblock In \emph{European Conference on Computer Vision}, 57--73. Springer.

\bibitem[{Liu et~al.(2022{\natexlab{c}})Liu, Li, Wu, Liu, Chen, Wu, and Li}]{automix}
Liu, Z.; Li, S.; Wu, D.; Liu, Z.; Chen, Z.; Wu, L.; and Li, S.~Z. 2022{\natexlab{c}}.
\newblock AutoMix: Unveiling the Power of Mixup for Stronger Classifiers.
\newblock In \emph{European Conference on Computer Vision}, 441--458.

\bibitem[{Montabone and Soto(2010)}]{montabone}
Montabone, S.; and Soto, A. 2010.
\newblock Human detection using a mobile platform and novel features derived from a visual saliency mechanism.
\newblock \emph{Image and Vision Computing}, 28(3): 391--402.

\bibitem[{Naseer et~al.(2021)Naseer, Ranasinghe, Khan, Hayat, Shahbaz~Khan, and Yang}]{naseer2021intriguing}
Naseer, M.~M.; Ranasinghe, K.; Khan, S.~H.; Hayat, M.; Shahbaz~Khan, F.; and Yang, M.-H. 2021.
\newblock Intriguing properties of vision transformers.
\newblock \emph{Advances in Neural Information Processing Systems}, 34: 23296--23308.

\bibitem[{Riche et~al.(2013)Riche, Duvinage, Mancas, Gosselin, and Dutoit}]{riche2013saliency}
Riche, N.; Duvinage, M.; Mancas, M.; Gosselin, B.; and Dutoit, T. 2013.
\newblock Saliency and human fixations: State-of-the-art and study of comparison metrics.
\newblock In \emph{Proceedings of the IEEE international conference on computer vision}, 1153--1160.

\bibitem[{Schwinn et~al.(2022)Schwinn, Precup, Eskofier, and Zanca}]{schwinn2022behind}
Schwinn, L.; Precup, D.; Eskofier, B.; and Zanca, D. 2022.
\newblock Behind the Machine’s Gaze: Neural Networks with Biologically-inspired Constraints Exhibit Human-like Visual Attention.
\newblock \emph{Transactions on Machine Learning Research}.

\bibitem[{Selvaraju et~al.(2017)Selvaraju, Cogswell, Das, Vedantam, Parikh, and Batra}]{cam}
Selvaraju, R.~R.; Cogswell, M.; Das, A.; Vedantam, R.; Parikh, D.; and Batra, D. 2017.
\newblock Grad-cam: Visual explanations from deep networks via gradient-based localization.
\newblock In \emph{Proceedings of the IEEE international conference on computer vision}, 618--626.

\bibitem[{Shi, Darrell, and Wang(2023)}]{absvit}
Shi, B.; Darrell, T.; and Wang, X. 2023.
\newblock Top-Down Visual Attention from Analysis by Synthesis.
\newblock In \emph{Proceedings of the IEEE/CVF Conference on Computer Vision and Pattern Recognition}, 2102--2112.

\bibitem[{Tan et~al.(2022)Tan, Gao, Wu, Li, and Li}]{Tan_2022_CVPR}
Tan, C.; Gao, Z.; Wu, L.; Li, S.; and Li, S.~Z. 2022.
\newblock Hyperspherical Consistency Regularization.
\newblock In \emph{Proceedings of the IEEE/CVF Conference on Computer Vision and Pattern Recognition (CVPR)}, 7244--7255.

\bibitem[{Touvron et~al.(2021{\natexlab{a}})Touvron, Cord, Douze, Massa, Sablayrolles, and J{\'e}gou}]{datahungry2}
Touvron, H.; Cord, M.; Douze, M.; Massa, F.; Sablayrolles, A.; and J{\'e}gou, H. 2021{\natexlab{a}}.
\newblock Training data-efficient image transformers \& distillation through attention.
\newblock In \emph{International conference on machine learning}, 10347--10357. PMLR.

\bibitem[{Touvron et~al.(2021{\natexlab{b}})Touvron, Cord, Douze, Massa, Sablayrolles, and J{\'e}gou}]{deit}
Touvron, H.; Cord, M.; Douze, M.; Massa, F.; Sablayrolles, A.; and J{\'e}gou, H. 2021{\natexlab{b}}.
\newblock Training data-efficient image transformers \& distillation through attention.
\newblock In \emph{International conference on machine learning}, 10347--10357. PMLR.

\bibitem[{Treisman and Gelade(1980)}]{treisman1980feature}
Treisman, A.~M.; and Gelade, G. 1980.
\newblock A feature-integration theory of attention.
\newblock \emph{Cognitive psychology}, 12(1): 97--136.

\bibitem[{Uddin et~al.(2021)Uddin, Monira, Shin, Chung, and Bae}]{saliencymix}
Uddin, A. F. M.~S.; Monira, M.~S.; Shin, W.; Chung, T.; and Bae, S.-H. 2021.
\newblock SaliencyMix: A Saliency Guided Data Augmentation Strategy for Better Regularization.
\newblock In \emph{International Conference on Learning Representations}.

\bibitem[{Verma et~al.(2019)Verma, Lamb, Beckham, Najafi, Mitliagkas, Lopez-Paz, and Bengio}]{mainfoldmix}
Verma, V.; Lamb, A.; Beckham, C.; Najafi, A.; Mitliagkas, I.; Lopez-Paz, D.; and Bengio, Y. 2019.
\newblock Manifold mixup: Better representations by interpolating hidden states.
\newblock In \emph{International conference on machine learning}, 6438--6447. PMLR.

\bibitem[{Vuyyuru et~al.(2020)Vuyyuru, Banburski, Pant, and Poggio}]{vuyyuru2020biologically}
Vuyyuru, M.~R.; Banburski, A.; Pant, N.; and Poggio, T. 2020.
\newblock Biologically inspired mechanisms for adversarial robustness.
\newblock \emph{Advances in Neural Information Processing Systems}, 33: 2135--2146.

\bibitem[{Wah et~al.(2011)Wah, Branson, Welinder, Perona, and Belongie}]{cub}
Wah, C.; Branson, S.; Welinder, P.; Perona, P.; and Belongie, S. 2011.
\newblock The caltech-ucsd birds-200-2011 dataset.

\bibitem[{Walawalkar et~al.(2020)Walawalkar, Shen, Liu, and Savvides}]{attentivecutmix}
Walawalkar, D.; Shen, Z.; Liu, Z.; and Savvides, M. 2020.
\newblock Attentive Cutmix: An Enhanced Data Augmentation Approach for Deep Learning Based Image Classification.
\newblock In \emph{ICASSP, IEEE International Conference on Acoustics, Speech and Signal Processing-Proceedings}.

\bibitem[{Wu et~al.(2023)Wu, Lin, Tan, Gao, and Li}]{9632431}
Wu, L.; Lin, H.; Tan, C.; Gao, Z.; and Li, S.~Z. 2023.
\newblock Self-Supervised Learning on Graphs: Contrastive, Generative, or Predictive.
\newblock \emph{IEEE Transactions on Knowledge and Data Engineering}, 35(4): 4216--4235.

\bibitem[{Xia et~al.(2022)Xia, Zhu, Du, and Li}]{xia2022pretraining}
Xia, J.; Zhu, Y.; Du, Y.; and Li, S.~Z. 2022.
\newblock Pre-training Graph Neural Networks for Molecular Representations: Retrospect and Prospect.
\newblock In \emph{ICML 2022 2nd AI for Science Workshop}.

\bibitem[{Xiao et~al.(2018)Xiao, Liu, Zhou, Jiang, and Sun}]{upernet}
Xiao, T.; Liu, Y.; Zhou, B.; Jiang, Y.; and Sun, J. 2018.
\newblock Unified perceptual parsing for scene understanding.
\newblock In \emph{Proceedings of the European conference on computer vision (ECCV)}, 418--434.

\bibitem[{Xie et~al.(2017)Xie, Girshick, Doll{\'a}r, Tu, and He}]{resnext}
Xie, S.; Girshick, R.; Doll{\'a}r, P.; Tu, Z.; and He, K. 2017.
\newblock Aggregated residual transformations for deep neural networks.
\newblock In \emph{Proceedings of the IEEE conference on computer vision and pattern recognition}, 1492--1500.

\bibitem[{Yarbus(2013)}]{yarbus2013eye}
Yarbus, A.~L. 2013.
\newblock \emph{Eye movements and vision}.
\newblock Springer.

\bibitem[{Yun et~al.(2019)Yun, Han, Oh, Chun, Choe, and Yoo}]{cutmix}
Yun, S.; Han, D.; Oh, S.~J.; Chun, S.; Choe, J.; and Yoo, Y. 2019.
\newblock CutMix: Regularization Strategy to Train Strong Classifiers With Localizable Features.
\newblock In \emph{Proceedings of the IEEE/CVF International Conference on Computer Vision (ICCV)}.

\bibitem[{Zang et~al.(2022)Zang, Li, Wu, Wang, Wang, Shang, Sun, Li, and Li}]{Dlme}
Zang, Z.; Li, S.; Wu, D.; Wang, G.; Wang, K.; Shang, L.; Sun, B.; Li, H.; and Li, S.~Z. 2022.
\newblock DLME: Deep Local-Flatness Manifold Embedding.
\newblock In Avidan, S.; Brostow, G.; Ciss{\'e}, M.; Farinella, G.~M.; and Hassner, T., eds., \emph{Computer Vision -- ECCV 2022}, 576--592. Cham: Springer Nature Switzerland.

\bibitem[{Zhang et~al.(2018)Zhang, Cisse, Dauphin, and Lopez-Paz}]{mixup}
Zhang, H.; Cisse, M.; Dauphin, Y.~N.; and Lopez-Paz, D. 2018.
\newblock mixup: Beyond Empirical Risk Minimization.
\newblock In \emph{International Conference on Learning Representations}.

\bibitem[{Zhao et~al.(2022)Zhao, Wu, Zhuang, Li, and Jia}]{zhao2022tracking}
Zhao, Z.; Wu, Z.; Zhuang, Y.; Li, B.; and Jia, J. 2022.
\newblock Tracking objects as pixel-wise distributions.
\newblock In \emph{European Conference on Computer Vision}, 76--94. Springer.

\bibitem[{Zhou et~al.(2017)Zhou, Zhao, Puig, Fidler, Barriuso, and Torralba}]{zhou2017scene}
Zhou, B.; Zhao, H.; Puig, X.; Fidler, S.; Barriuso, A.; and Torralba, A. 2017.
\newblock Scene parsing through ade20k dataset.
\newblock In \emph{Proceedings of the IEEE conference on computer vision and pattern recognition}, 633--641.

\end{thebibliography}

\end{document}